\documentclass[sigconf]{acmart}
\usepackage{booktabs}
\usepackage{multirow}

\setcopyright{none}
\renewcommand\footnotetextcopyrightpermission[1]{}
\acmConference[]{}{}{}

\newcommand{\sys}{ReForesight}

\begin{document}

\title[\sys{}: Re-Identification and Future Occupancy Forecasting]{\sys{}: Occlusion-Robust Re-Identification and Future Occupancy Forecasting for Dynamic Semantic Mapping}

%% Author block (hidden in review mode by the "anonymous" option).
\author{Didar Seyidov}
\email{Didar.Seyidov@skoltech.ru}
\affiliation{%
  \institution{Skolkovo Institute of Science and Technology}
  \city{Moscow}
  \country{Russia}}

\author{Faith Ogunwoye}
\email{Faith.Ogunwoye@skoltech.ru}
\affiliation{%
  \institution{Skolkovo Institute of Science and Technology}
  \city{Moscow}
  \country{Russia}}

\author{Hajira Amjad}
\email{Hajira.Amjad@skoltech.ru}
\affiliation{%
  \institution{Skolkovo Institute of Science and Technology}
  \city{Moscow}
  \country{Russia}}

\author{Danil Tolkachev}
\email{tolkachev@168robotics.com}
\affiliation{%
  \institution{168robotics}
  \city{Moscow}
  \country{Russia}}

\author{Yuri Sarkisov}
\email{ysarkisov@168robotics.com}
\affiliation{%
  \institution{168robotics}
  \city{Moscow}
  \country{Russia}}

\author{Dzmitry Tsetserukou}
\email{D.Tsetserukou@skoltech.ru}
\affiliation{%
  \institution{Skolkovo Institute of Science and Technology}
  \city{Moscow}
  \country{Russia}}

\begin{abstract}
Robots sharing walkways need to know where people will be, yet semantic occupancy maps show only where they are. We present \sys{}, an open-source ROS~2 toolkit that adds instance-level future occupancy and re-identification after occlusion to a particle-based semantic mapping pipeline. A 36.6k-parameter MLP predicts each tracked person or vehicle up to 3\,s ahead, guarded by a Kalman fallback and a stationary gate, and the object's own points are moved to the forecast pose, forming a ``ghost'' point cloud in the map frame without an occupancy decoder. An optional tracker re-associates objects after occlusion using keypoint descriptors the pipeline already computes and a metric 3D gate. \textbf{On four held-out CODa sequences, the MLP lowers the mean 3\,s final displacement error from 0.728\,m (Kalman) to 0.662\,m and beats a similar-size GRU (0.692\,m) on every sequence, at 0.04\,ms per instance on one CPU core.} Code, all models, and a ROS-free demo are released; field demonstrations use an outdoor cleaning robot.
\end{abstract}




\begin{teaserfigure}
  \centering
  \includegraphics[width=\textwidth]{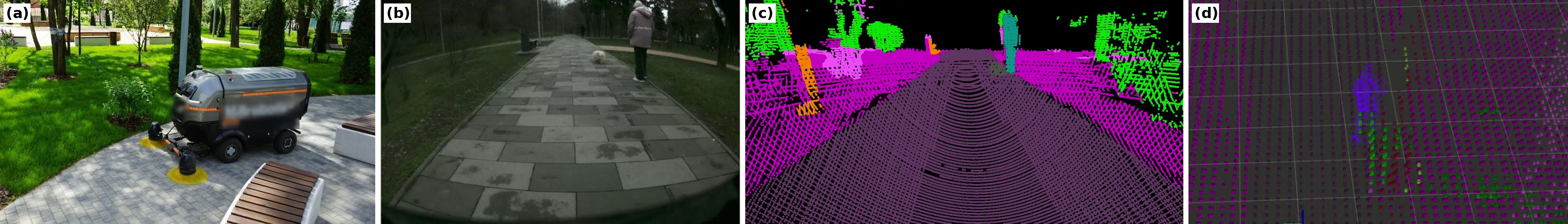}
  \caption{\sys{} on the field robot. (a)~Outdoor cleaning robot used for the recordings (logo removed for review). (b)~Front-camera view in a public park. (c)~Semantic occupancy map projected into the camera (teal: pedestrian instance). (d)~The robot's 3D semantic occupancy map with the pedestrian's forecast occupancy 1\,s ahead (indigo ``ghost'').}
  \Description{Four panels. Panel a shows a gray wheeled cleaning robot with side brushes on a paved park path. Panel b shows a fisheye camera view of a park walkway with a pedestrian and a small dog. Panel c shows the same view rendered as colored voxels, with the pedestrian in teal, the path in purple and trees in green. Panel d shows a 3D voxel map in purple and green with a pedestrian-shaped cluster of indigo points ahead of the robot.}
  \label{fig:teaser}
\end{teaserfigure}

\maketitle

\section{Introduction}
In recent years, robots have increasingly been operating in close proximity to pedestrians on walkways, campuses, and in parks. Navigating such spaces safely and in a socially acceptable way requires anticipating people's behavior and movements, not only their current positions~\cite{kruse2013survey,mavrogiannis2023core,rudenko2020survey}; a robot that cannot predict human motion tends to react late or to stop altogether~\cite{trautman2010unfreezing}, increasing the risk of unsafe interactions, collisions, and harm to people nearby. Existing work addresses separate parts of this problem. Instance-aware semantic occupancy maps~\cite{chen2025dspmap} and dynamic-object mapping systems~\cite{schmid2023dynablox,schmid2024khronos} represent moving objects in a metric map but expose only their current or past state. Trajectory predictors~\cite{alahi2016sociallstm,salzmann2020trajectron,bae2024singular} forecast object centroids, and prediction-aware crowd navigation~\cite{samavi2025sicnav} consumes such trajectories, but neither produces occupancy in the robot's map. Learned 4D occupancy forecasting~\cite{hu2021fiery,khurana2023pcf,ma2024cam4docc,zheng2024occworld} does predict future occupancy, but with large neural networks trained for driving setups, which are hard to run deterministically and with low latency on the power-efficient compact computers used in service robots.

\sys{} fills this gap with a deliberately lightweight design built on one observation: a tracked instance in the map already carries its own observed geometry, so moving that geometry along a forecast trajectory yields instance-shaped future occupancy directly. The contribution is therefore not a new occupancy network but the integration of instance forecasting into a semantic map that preserves each object's real extent and identity and needs no occupancy decoder. \sys{} subscribes to topics the mapping pipeline already publishes, forecasts each tracked instance with a compact learned model guarded by two safety gates, and re-associates objects after occlusion, which the forecaster needs because it relies on uninterrupted track histories. We release it as open-source software with three main components:
\begin{itemize}
  \item \textbf{Instance-level future occupancy.} A ROS~2~\cite{macenski2022ros2} node that lifts each tracked instance into the map frame, forecasts it with a compact learned model (a 36,594-parameter MLP by default; a 29,787-parameter GRU is included for comparison), and publishes forecast trajectories and ``ghost'' point clouds built from the instance's own observed points.
  \item \textbf{Re-identification after occlusion.} A two-stage tracker, \texttt{ReIDTracker}, re-associates objects that disappear for up to 3\,s (verified in synthetic tests). Its appearance cue pools SuperPoint descriptors that the pipeline already computes, so it adds no network inference, and its metric 3D motion gate resolves depth-aliased candidates that a pixel gate confuses.
  \item \textbf{A lightweight, reproducible release.} CPU forecasting that leaves the GPU to the perception networks, a ROS~2 port of the Semantic DSP Map, CODa replay tools, all trained checkpoints, unit tests, a Docker image, and a ROS-free demo that reproduces every table in this paper.
\end{itemize}

\begin{figure*}[t]
  \centering
  \includegraphics[width=0.92\textwidth]{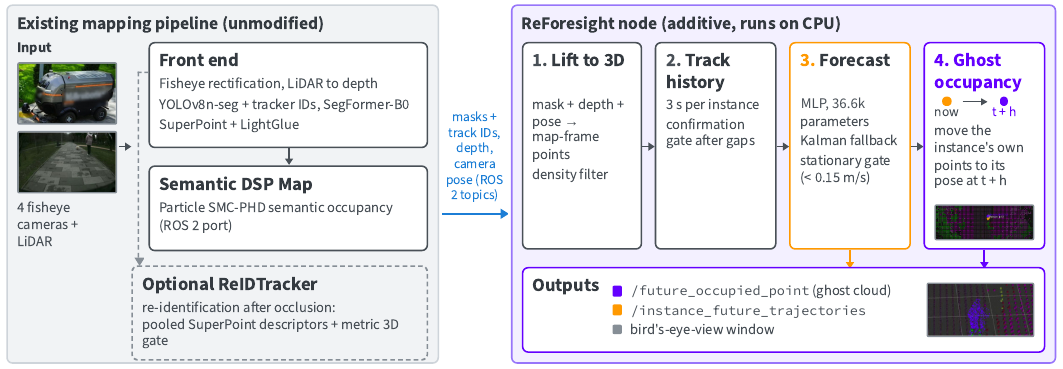}
  \caption{\sys{} architecture. The unmodified mapping pipeline (left) publishes instance masks with track IDs, depth, and camera pose. The \sys{} node (right) lifts each instance into the map frame, keeps a 3\,s pose history, forecasts it with the learned forecaster (MLP by default, or the Kalman fallback), applies the confirmation and stationary gates, and moves the instance's own points to the forecast pose. Dashed: optional re-identification module.}
  \Description{Block diagram in two panels. Left gray panel, existing mapping pipeline: input thumbnails of the robot and its front camera feed a front end (fisheye rectification, LiDAR to depth, YOLOv8n-seg with tracker IDs, SegFormer-B0, SuperPoint and LightGlue) and the Semantic DSP Map; a dashed optional ReIDTracker block sits below. A blue arrow labeled with the ROS 2 topics (masks with track IDs, depth, camera pose) leads to the right violet panel, the ReForesight node, with four blocks in a row: lift to 3D, track history, forecast (MLP, Kalman fallback, stationary gate), and ghost occupancy with a bird's-eye-view thumbnail. Below them an outputs block lists the ghost point cloud topic, the trajectory topic, and the bird's-eye-view window, next to a 3D thumbnail of an indigo ghost pedestrian.}
  \label{fig:pipeline}
\end{figure*}

\section{Proposed Solution Architecture}
\label{sec:system}
Figure~\ref{fig:pipeline} summarizes the architecture. The forecasting node is purely additive: it does not modify or recompile the mapping core, and its launch file includes the original pipeline launch file unchanged.

\textbf{Mapping pipeline.} The Semantic DSP Map~\cite{chen2025dspmap} estimates occupancy, semantics, and instance IDs with a state-augmented sequential Monte Carlo PHD filter and was released for ROS~1. We ported it to ROS~2 and built a front end for our robot: fisheye images are rectified to a shared pinhole model, the merged LiDAR cloud is projected into a depth image, YOLOv8n-seg~\cite{jocher2023yolov8} provides instance masks with IDs from its default BoT-SORT tracker~\cite{aharon2022botsort}, SegFormer-B0~\cite{xie2021segformer} labels the background, and SuperPoint~\cite{detone2018superpoint} with LightGlue~\cite{lindenberger2023lightglue} matches keypoints for per-instance motion estimation. The camera pose is stamped with the depth image's time stamp so that depth and pose refer to the same LiDAR sweep. An experimental multiplexer feeds up to four cameras into one map.

\textbf{Track histories.} The node synchronizes each mask group with the depth image and camera pose. For every dynamic instance it back-projects the mask pixels with valid depth into the map frame, removes sparse silhouette-edge points with a voxel density filter (0.12\,m voxels, at least 5 points in the $3{\times}3{\times}3$ neighborhood), keeps at most 2,000 points, and appends the instance position to a 3\,s history. Heading follows the displacement direction once an object moves more than 5\,cm per step. Tracks unseen for 2\,s are dropped. After an observation gap longer than 0.15\,s, a track must be observed for two further consecutive frames before it is forecast again, which protects the forecast from a single wrong association after an occlusion.

\textbf{Forecaster.} The default MLP reads 19 relative increments $(\Delta x,\allowbreak \Delta y,\allowbreak \Delta\psi)$ from a 2\,s history, expressed in the object's heading frame, with an 8-dimensional embedding of the motion class (pedestrian, vehicle, bicycle), and maps them through two hidden layers of 128 units to 30 future increments (3\,s at 10\,Hz). For comparison we also train a GRU~\cite{cho2014gru} encoder (hidden size 64) with an autoregressive GRU-cell decoder on the same inputs. Both are exported to TorchScript and are interchangeable in the node. Training minimizes $\mathrm{ADE} + 0.5\,\mathrm{FDE}$ plus a small heading term, uses scheduled sampling~\cite{bengio2015scheduled} for the GRU, and draws windows with weights balanced over class and motion regime (moving, stationary, accelerating, decelerating), because CODa is dominated by continuously walking pedestrians (Adam, learning rate $10^{-3}$, batch 256, early stopping on validation ADE). Tracks with fewer than 20 samples, or systems without PyTorch, use a constant-velocity Kalman filter~\cite{kalman1960}. A stationary gate replaces the forecast with the current pose when the object's measured net speed over the last 1\,s is below 0.15\,m/s. We added it after synthetic stopped histories showed that the learned model keeps extrapolating motion for people who have just stopped; the threshold was set from detection jitter and walking speed, not tuned on any test sequence.

\textbf{Future occupancy.} For a horizon $h$ of 0.5, 1, 2, or 3\,s, the instance's current point cluster is rotated about its centroid by the predicted heading change and translated to the predicted position; heights are unchanged. The 3\,s maximum equals the training horizon and CODa's 3\,s instance-persistence rule. The 1\,s default keeps the expected error on CODa (0.234\,m for the MLP with the gate, Table~\ref{tab:coda}) close to the robot map's 0.2\,m voxel size while still covering about 1.3\,m of walking at typical pedestrian speed. The ghost clouds of all instances are published as one \texttt{PointCloud2} in the map frame, with \texttt{MarkerArray} trajectories; because the shape comes from the latest observation, each ghost keeps the size and extent of the real object.

\textbf{Optional re-identification.} The front end's default tracker associates detections by motion and overlap only. \texttt{ReIDTracker} adds a second stage that re-identifies tracks lost for at most 30 frames (CODa's instance persistence rule) by the cosine similarity of SuperPoint descriptors mean-pooled inside each mask. Its motion gate operates in metric 3D when depth is available (radius 0.5\,m, growing by 0.35\,m per lost frame) and in pixels otherwise, and a re-identified track stays unconfirmed for two frames. It ships as a unit-tested patch to the front end and is disabled by default.

\begin{figure*}[t]
  \centering
  \includegraphics[width=\textwidth]{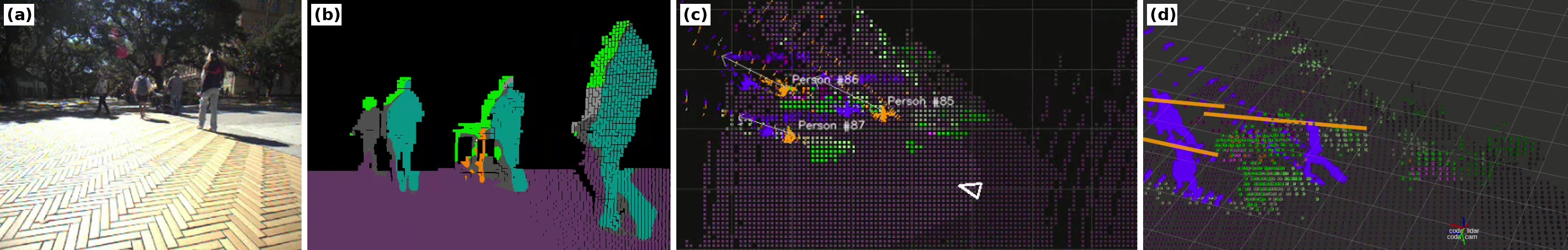}
  \caption{\sys{} replaying the public CODa dataset (one time step of sequence 20). (a)~Camera image. (b)~Semantic occupancy map projected into the camera (teal: pedestrian instances). (c)~Bird's-eye view: current pedestrian points (orange), ghosts 1\,s ahead (indigo), gray arrows from each current centroid to its ghost, robot at the white triangle. (d)~3D map with the ghosts (indigo) and predicted 3\,s trajectories (orange).}
  \Description{Four panels from a university campus walkway. Panel a is a sunny camera image with several pedestrians. Panel b shows the pedestrians as teal voxels on a purple ground surface. Panel c is a top-down dark view with three orange point clusters labeled Person 85, 86 and 87, each connected by a gray arrow to an indigo cluster displaced along the walking direction. Panel d is a 3D voxel map with three indigo pedestrian-shaped clusters and orange line segments showing predicted paths.}
  \label{fig:qual}
\end{figure*}

\section{Evaluation}
\label{sec:eval}

\textbf{Forecasting on CODa.} CODa~\cite{zhang2024coda} provides 3D boxes with persistent instance IDs at 10\,Hz. We convert them into world-frame trajectories using CODa's globally aligned poses, split tracks at gaps longer than 30 frames or implied speeds above 36\,m/s (reused IDs), fill short gaps linearly, and cut windows of 2\,s history and 3\,s future with a 0.5\,s stride, merging vehicle-like classes as CODa does. The main split trains on seven sequences (32,267 windows), validates on sequence 20, and tests on sequence 7 (5,812 windows, 184 tracks). For cross-sequence evaluation we also hold out sequences 13, 16, and 3 (256, 217, and 42 tracks), each time training on the remaining training sequences. Every learned model is trained with three seeds, and the run with the lowest validation ADE is reported; the released checkpoints are these selected runs. Baselines are the constant-velocity Kalman filter that also serves as the run-time fallback, and a raw constant-velocity model from the last observed step.

\begin{table}[t]
  \caption{Displacement error (m) by prediction horizon on the held-out CODa sequence 7 (5,812 windows, 184 tracks). The 3\,s column is the FDE. Best value per column in bold.}
  \label{tab:coda}
  \small
  \setlength{\tabcolsep}{4.5pt}
  \begin{tabular}{@{}lccccc@{}}
    \toprule
    & \multicolumn{4}{c}{Error at horizon} & \\
    \cmidrule(lr){2-5}
    Method & 0.5\,s & 1\,s & 2\,s & 3\,s & ADE \\
    \midrule
    \multicolumn{6}{@{}l}{\emph{Without stationary gate}} \\
    \quad CV (last step) & 0.337 & 0.643 & 1.262 & 1.899 & 0.986 \\
    \quad CV Kalman & 0.165 & 0.264 & 0.481 & 0.708 & 0.387 \\
    \quad MLP       & 0.153 & 0.236 & 0.429 & 0.660 & 0.351 \\
    \quad GRU & 0.164 & 0.256 & 0.446 & 0.664 & 0.365 \\
    \midrule
    \multicolumn{6}{@{}l}{\emph{With stationary gate}} \\
    \quad CV Kalman & 0.163 & 0.260 & 0.472 & 0.694 & 0.380 \\
    \quad MLP (released default) & \textbf{0.152} & \textbf{0.234} & \textbf{0.423} & \textbf{0.648} & \textbf{0.346} \\
    \quad GRU & 0.163 & 0.254 & 0.444 & 0.658 & 0.363 \\
    \bottomrule
  \end{tabular}
\end{table}

\begin{table}[t]
  \caption{Cross-sequence 3\,s FDE (m): each CODa sequence held out in turn (test tracks in parentheses). Learned models: lowest-validation-ADE run of three seeds. Best per row and gate setting in bold.}
  \label{tab:cv}
  \small
  \setlength{\tabcolsep}{3.4pt}
  \begin{tabular}{@{}lcccccc@{}}
    \toprule
    & \multicolumn{3}{c}{Without gate} & \multicolumn{3}{c}{With gate} \\
    \cmidrule(lr){2-4}\cmidrule(l){5-7}
    Held-out seq. & Kalman & MLP & GRU & Kalman & MLP & GRU \\
    \midrule
    7 (184)  & 0.708 & \textbf{0.660} & 0.664 & 0.694 & \textbf{0.648} & 0.658 \\
    13 (256) & 0.895 & \textbf{0.892} & 0.949 & 0.891 & \textbf{0.889} & 0.945 \\
    16 (217) & 0.832 & \textbf{0.731} & 0.772 & 0.813 & \textbf{0.720} & 0.735 \\
    3 (42)   & 0.475 & \textbf{0.367} & 0.381 & 0.466 & \textbf{0.356} & 0.375 \\
    \midrule
    Mean     & 0.728 & \textbf{0.662} & 0.692 & 0.716 & \textbf{0.653} & 0.678 \\
    \bottomrule
  \end{tabular}
\end{table}

Table~\ref{tab:coda} reports the main split. \textbf{\boldmath Both learned models outperform the Kalman filter at every horizon; the MLP lowers FDE by 6.9\% (0.708 to 0.660\,m), with a track-level 95\% bootstrap interval of the difference of $[-0.079,\allowbreak -0.018]$\,m, and remains significantly better with the gate applied to both ($[-0.076,\allowbreak -0.015]$\,m).} The raw constant-velocity model is far worse (1.899\,m) because single-step velocities amplify 10\,Hz annotation jitter, so the filtered Kalman model is the relevant baseline. Table~\ref{tab:cv} shows how this generalizes. \textbf{Across the four held-out sequences, the MLP lowers mean FDE by 9.0\% relative to the Kalman filter, with gains of up to 22.8\% on sequence 3, and is more accurate than the GRU on every sequence.} The GRU lowers mean FDE by 5.0\% but is 6.0\% worse than the Kalman filter on sequence 13, where the MLP still improves slightly (0.4\%). The MLP is also less sensitive to the training seed (FDE 0.660 to 0.691\,m on sequence 7, against 0.664 to 0.772\,m for the GRU) and, with the gate applied to both, only the MLP remains significantly better than the Kalman filter (GRU interval $[-0.074,\allowbreak 0.006]$\,m). \textbf{The stationary gate lowers FDE for every method on every held-out sequence}; on the stationary windows of sequence 7 (14.2\%) it reduces the MLP's FDE by 38\% (0.215 to 0.134\,m). Pedestrian gains are significant, while vehicle and bicycle results are inconclusive with only 20 and 16 test tracks in sequence 7. We therefore release the MLP as the default forecaster and keep the GRU for comparison.

\textbf{Run time and portability.} Timings come from a desktop (Intel i9-14900KF, NVIDIA RTX 5090) replaying recorded data. We report single-thread medians on an efficiency core, which is closer to the low-power cores of embedded computers than the desktop's performance cores: the TorchScript MLP takes 0.04\,ms per instance (0.08\,ms for a batch of 20) and the GRU 1.92\,ms (3.15\,ms). The mapping core logged median update times of 0.350\,s on the robot recordings ($n{=}366$) and 0.214\,s on CODa ($n{=}8{,}173$), so forecasting is not the bottleneck. The node needs only three ROS~2 topics (instance masks with IDs, depth, camera pose) and a CPU; on the robot, the pipeline runs on its onboard NVIDIA Jetson computer, for which we provide a headless deployment configuration.

\textbf{Field demonstrations.} We recorded two runs of about 2\,min each with a commercial outdoor cleaning robot in a public park (four fisheye cameras and a merged LiDAR at about 10\,Hz; we use the front camera). Figure~\ref{fig:teaser}(b to d) shows a pedestrian and its ghost 1\,s ahead. These recordings have no ground truth, so they demonstrate pipeline operation, not forecast accuracy (Figures~\ref{fig:teaser} and~\ref{fig:qual} were recorded with the GRU checkpoint). Figure~\ref{fig:qual} shows the pipeline replaying CODa sequence 20 (validation, not used for gradient updates).

\begin{table}[t]
  \caption{\texttt{ReIDTracker} ablation on synthetic occlusion scenarios (100 seeds each). S1: identity kept after 1 to 30 missed frames / not re-used after 35 to 45. S2: a different object enters an occluded track's gate and should get a new identity. S3: a distant decoy projects onto the predicted pixel location. S4: two objects cross (mean identity changes per run). S1, S2, and S4 use 3D positions back-projected at 8\,m depth. Best in bold.}
  \label{tab:reid}
  \small
  \setlength{\tabcolsep}{3.0pt}
  \begin{tabular}{@{}lccccc@{}}
    \toprule
    & \multicolumn{2}{c}{S1 occlusion (\%)} & S2 (\%) & S3 (\%) & S4 \\
    \cmidrule(lr){2-3}
    Variant & keep & refuse & new ID & decoy & changes \\
    \midrule
    IoU only (no memory)      & 0.0            & \textbf{100.0} & \textbf{100.0} & 0.0            & \textbf{0.54} \\
    Motion, 2D gate           & \textbf{100.0} & \textbf{100.0} & 0.0            & 0.0            & \textbf{0.54} \\
    Motion, 3D gate           & \textbf{100.0} & \textbf{100.0} & 0.0            & \textbf{100.0} & \textbf{0.54} \\
    Appearance, 2D gate       & \textbf{100.0} & \textbf{100.0} & \textbf{100.0} & \textbf{100.0} & \textbf{0.54} \\
    Appearance, 3D gate (full)& \textbf{100.0} & \textbf{100.0} & \textbf{100.0} & \textbf{100.0} & \textbf{0.54} \\
    \bottomrule
  \end{tabular}
\end{table}

\textbf{Re-identification.} Table~\ref{tab:reid} ablates \texttt{ReIDTracker} on scripted synthetic scenarios included in the release. \textbf{With appearance, the tracker succeeds in all of S1 to S3: keeping lost tracks recovers every occlusion of up to 30 frames, appearance rejects a different object entering the occluded track's gate (100\% versus 0\% for motion alone), and either appearance or the metric 3D gate avoids the depth-aliased decoy.} The 3D gate therefore matters when descriptors are unavailable, for example on low-texture objects. The IoU-only tracker passes S2 trivially because it never re-identifies. S4 shows a limitation shared by all variants: in 27 of 100 seeds, two crossing objects swap identities, because the first, overlap-based stage does not consult appearance. Our claim is therefore limited to re-association after disappearance; the module is not yet enabled by default, and an evaluation against BoT-SORT on annotated real data is future work.

\section{Release}
\label{sec:release}
\textbf{Contents and license.} The supplementary archive contains the \sys{} package, the ROS~2 port of the Semantic DSP Map, launch files for the robot and CODa, all trained checkpoints (released MLP and GRU plus every run behind Tables~\ref{tab:coda} and~\ref{tab:cv}), and the CODa-derived trajectory windows. \sys{} builds directly on the open-source Semantic DSP Map~\cite{chen2025dspmap}, and we thank its authors for releasing it. Code is released under Apache-2.0, the license of the upstream mapper; the trajectory windows inherit CODa's CC BY-NC-SA 4.0 license. At camera-ready the code will move to a public repository with an issue tracker and versioned releases archived on Zenodo with a DOI.

\textbf{Platform and demo.} The toolkit targets Ubuntu 24.04 with ROS~2 Jazzy and, for on-robot use, NVIDIA Jetson with ROS~2 Humble. A Dockerfile and a single script run the unit tests and a ROS-free demo that reproduces Tables~\ref{tab:coda} to~\ref{tab:reid} in under a minute on a CPU; one launch command replays a raw CODa sequence through the full pipeline.

\textbf{Intended uses.} (i)~Human-aware navigation: the ghost cloud is a standard \texttt{PointCloud2} in the map frame, so costmap layers and planners that consume point clouds can react to where people are about to be. (ii)~Studying and communicating anticipation: ghosts and the bird's-eye view make the robot's predictions visible to experimenters and could be shown to participants in studies of robot legibility. (iii)~Reproducible experiments: other predictors can be swapped in and compared within the same perception stack on public data.

\textbf{Responsible use and limitations.} Forecasts are deterministic point estimates learned from one university campus; they are not safety certified and must not be a robot's only safety layer. The models are unimodal and the propagation is planar and rigid, which ignores articulated motion and agent interactions. We have not compared against heavy 4D occupancy networks on the same hardware, evaluated closed-loop navigation, run a user study, or tested re-identification on real data, and the live front end still labels cyclists as people. Camera recordings in public spaces contain bystanders; the archive contains no raw robot images, and any future release will blur faces and follow data-protection rules.

\section{Conclusion}
\sys{} turns an instance-aware semantic map into a forecast of where nearby people will be, publishes it as standard ROS~2 topics in the robot's map frame, and re-associates people after occlusion with descriptors the pipeline already computes. Across four held-out CODa sequences, the released 36.6k-parameter MLP lowers the mean 3\,s FDE of a constant-velocity Kalman filter from 0.728\,m to 0.662\,m (9.0\%), outperforms a similar-size GRU on every sequence, and costs 0.04\,ms per instance on a CPU efficiency core; the stationary gate improves every method. Next steps are multimodal forecasts, comparisons with heavier occupancy forecasters, closed-loop navigation, and user studies of visualized ghosts.

\bibliographystyle{ACM-Reference-Format}
\bibliography{refs}

\end{document}